\documentclass[pmlr]{jmlr}

\usepackage{bm}
\usepackage{color}
\newcommand{\RR}{\mathbb{R}}
\newcommand{\CC}{\mathbb{C}}
\newcommand{\dd}{\mathrm{d}}
\renewcommand{\SS}{\mathbb{S}}

\newcommand{\calS}{\mathcal{S}}
\newcommand{\calH}{\mathcal{H}}
\newcommand{\calA}{\mathcal{A}}
\newcommand{\calM}{\mathcal{M}}

\renewcommand{\aa}{\bm{a}}
\newcommand{\xx}{\bm{x}}
\newcommand{\yy}{\bm{y}}
\renewcommand{\tt}{\bm{t}}
\newcommand{\uu}{\bm{u}}
\newcommand{\xxi}{\bm{\xi}}

\newcommand{\tI}{\tilde{I}}
\newcommand{\tx}{\tilde{\xx}}
\newcommand{\ta}{\tilde{\aa}}
\newcommand{\tL}{\tilde{L}}

\newcommand{\nn}{\mathtt{NN}}
\newcommand{\ridge}{\mathtt{R}}

\DeclareMathOperator{\aff}{Aff}
\DeclareMathOperator{\id}{Id}

\newcommand{\pihat}{\widehat{\pi}}

\newcommand{\iprod}[1]{\langle#1\rangle}
\newcommand{\iiprod}[1]{(\!(#1)\!)}

\newcommand{\refeq}[1]{\eqref{eq:#1}}
\newcommand{\reffig}[1]{Figure~\ref{fig:#1}}

\newcommand{\refapp}[1]{Appendix~\ref{sec:#1}}

\newcommand{\refthm}[1]{Theorem~\ref{thm:#1}}
\newcommand{\reflem}[1]{Lemma~\ref{lem:#1}}

\jmlrvolume{}
\firstpageno{1}
\editors{Sophia Sanborn, Christian Shewmake, Simone Azeglio, Nina Miolane}

\jmlryear{2023}
\jmlrworkshop{Symmetry and Geometry in Neural Representations}

\title[Joint Invariants on Data-Parameter Domain Induce Universal Neural Networks]{Joint Group Invariant Functions on Data-Parameter Domain Induce Universal Neural Networks}

\author{%
\Name{Sho Sonoda}${}^1$ \Email{sho.sonoda@riken.jp}\\
\Name{Hideyuki Ishi}${}^2$ \Email{hideyuki-ishi@omu.ac.jp}\\
\Name{Isao Ishikawa}${}^{3,1}$ \Email{ishikawa.isao.zx@ehime-u.ac.jp}\\
\Name{Masahiro Ikeda}${}^1$ \Email{masahiro.ikeda@riken.jp}\\
\addr ${}^1$Center for Advanced Intelligence Project (AIP), RIKEN\\
${}^2$Osaka Central Advanced Mathematical Institute (OCAMI), Osaka Metropolitan University\\
${}^3$Center for Data Science, Ehime University%
}

\begin{document}

\maketitle

\begin{abstract}
    The symmetry and geometry of input data are considered to be encoded in the internal data representation inside the neural network, but the specific encoding rule has been less investigated. In this study, we present a systematic method to induce a generalized neural network and its right inverse operator, called the \emph{ridgelet transform}, from a \emph{joint group invariant function} on the data-parameter domain. Since the ridgelet transform is an inverse, (1) it can describe the arrangement of parameters for the network to represent a target function, which is understood as the \emph{encoding rule}, and (2) it implies the \emph{universality} of the network. Based on the group representation theory, we present a new simple proof of the universality by using Schur's lemma in a unified manner covering a wide class of networks, for example, the original ridgelet transform, formal \emph{deep} networks, and the dual voice transform. Since traditional universality theorems were demonstrated based on functional analysis, this study sheds light on the group theoretic aspect of the approximation theory, connecting geometric deep learning to abstract harmonic analysis.
\end{abstract}


\section{Introduction}
The internal data representation of neural networks is expected to reflect the symmetry and geometry of the data domain. In geometric deep learning \citep{Bronstein2021}, several authors have developed novel network architectures that are compatible with the geometric structure of the data (e.g. group equivariant networks). However, these methods typically require handcrafting the network architecture for each specific symmetry and geometry.
In this study, we present a systematic method to induce a generalized neural network and its right inverse operator, called the \emph{ridgelet transform}, from a \emph{joint group invariant function} on the data-parameter domain.
Since the ridgelet transform is an inverse, (1) it explicitly describes the arrangement of parameters for the network to represent a target function, %
and (2) it implies the \emph{universality} of the network. 

\begin{remark}
Our reviewers have kindly let us know that \citet{Cohen2019}, \citet{Finzi2021}, and \citet{Aslan2023} have proposed versatile group equivariant network architectures that cover a wide class of groups in a unified manner, and \citet{Ravanbakhsh2017} have investigated the symmetry in the parameters. Since our results are applicable to \emph{any} network architectures, it would be interesting to find the ridgelet transform for each network.
\end{remark}

The proof of a universality theorem contains hints for understanding the internal data processing mechanisms inside neural networks. The year 1989 was the beginning of the universality theorem and a great year, as four different proofs were presented by \citet{Cybenko1989}, \citet{Hornik1989}, \citet{Funahashi1989}, and \citet{Carroll.Dickinson}. Among them, Cybenko's proof using Hahn-Banach and Hornik et al.'s proof using Stone-Weierstrass are existential proofs, meaning that it is not clear how to assign the parameters. On the other hand, Funahashi's proof reducing to the Fourier transform and Carroll and Dickinson's proof reducing to the Radon transform are constructive proofs, meaning that it is clear how to assign the parameters. The latter constructive methods, which reduce to integral transforms, were refined as the so-called integral representation by \citet{Barron1993} and further culminated as the ridgelet transform discovered by \citet{Murata1996} and \citet{Candes.PhD}.

The ridgelet transform, the main topic of this study, is a pseudo-inverse operator of the integral representation neural network and is a detailed analysis tool that can describe the relationship between data and parameters due to its analytical representation.
In the 2000s, thanks to the efforts of Donoho and others, research on ridgelet transforms evolved into geometric multiscale analysis \citep[GMA, see e.g.][]{Donoho2002}, leading to the development of various x-lets such as curvelets \citep{Candes2004curvelet}, contourlet \citep{Do2005contourlet}, shearlet \citep{Labate2005shearlet}, bandelet \citep{Pennec2005bandelet}, and grouplet \citep{Mallat2009grouplet}. These lines of studies mainly focused on developing multidimensional wavelet transforms for image processing (i.e., 2D signals) \citep{Starck2010,Mallat2009book} and gradually moved apart from neural networks.

In the 2020s, the concept of integral representations has re-emerged as tools for analyzing deep learning theories, bringing renewed attention to ridgelet transforms. Precisely, they are often referred to by different names such as overparametrization, continuous/infinite width, mean field theory \citep{Nitanda2017,Mei2018,Rotskoff2018,Chizat2018,Sirignano2020}, and Langevin dynamics \citep{Suzuki2020}. \citet{sonoda2022symmetric,Sonoda2022gconv} have developed ridgelet transforms for various networks, such as group convolutional networks and networks on manifolds, and have shown constructive universality theorems. In these proofs, reducing the network to Fourier transforms was an essential step to find the ridgelet transforms. In this study, we can find the ridgelet transforms even when there is no clear path to reducing them to Fourier transforms, as long as we can find a group invariant function. %

The theory of function expansion based on group representations is well investigated in abstract harmonic analysis \citep{Folland2015}. There are two main streams: one is the generalization of \emph{Fourier transform}, which expands functions on group $G$ as a sum/integration of multiple irreducible unitary representations \citep{Sugiura1990book}, and the other is the generalization of \emph{wavelet transform} called the \emph{voice transform}, which expands functions in representation space $\calH$ as a sum/integration of functions generated by a single square-integrable unitary representation \citep{Holschneider1998book,Berge2021}. 
For example, recent studies by \citet{miyato2022unsupervised} and \citet{Koyama2023} belong to the Fourier stream, while this study belongs to the wavelet/voice stream. Yet, it is precisely a new integral transform that differs from the conventional voice transform. 
The generalized ridgelet transform discovered in this study was motivated by the research objective of geometrically analyzing the parameters of neural networks, and we believe it is a missing link for connecting geometric deep learning to abstract harmonic analysis.

\section{Preliminaries}
We showcase the original integral representation and the ridgelet transform, a mathematical model of depth-2 fully-connected network and its right inverse, then list a few facts in the group representation theory.

\paragraph{Notation.}
For any topological space $X$, $C_c(X)$ denotes the Banach space of all compactly supported functions $f$ on $X$.
$\calS(\RR^d)$ and $\calS'(\RR^d)$ denote the classes of rapidly decreasing functions (or Schwartz test functions) and tempered distributions on $\RR^d$, respectively. 

\subsection{Quick Introduction to Integral Representation and Ridgelet Transform}

\begin{definition} For any measurable function $\sigma:\RR\to\CC$ and Borel measure $\gamma$ on $\RR^m\times\RR$, put
\begin{align}
    S_\sigma[\gamma](\xx) := \int_{\RR^m\times\RR} \gamma(\aa,b)\sigma(\aa\cdot\xx-b)\dd\aa\dd b, \quad \xx \in \RR^m.
\end{align}
We call
$S_\sigma[\gamma]$ an (integral representation of) neural network, and $\gamma$ a parameter distribution.
\end{definition}
The integration over all the hidden parameters $(\aa,b) \in \RR^m\times\RR$ means all the neurons $\{ \xx\mapsto\sigma(\aa\cdot\xx-b) \mid (\aa,b) \in \RR^m\times\RR \}$ are summed (or integrated, to be precise) with weight $\gamma$, hence formally $S_\sigma[\gamma]$ is understood as a continuous neural network with a single hidden layer.
We note, however, when $\gamma$ is a finite sum of point measures such as $\gamma_p = \sum_{i=1}^p c_i \delta_{(\aa_i,b_i)}$, then it can also reproduce a finite width network
\begin{equation}
    S_\sigma[\gamma_p](\xx) = \sum_{i=1}^p c_i \sigma(\aa_i\cdot\xx-b_i).
\end{equation}
In other words, the integral representation is a mathmatical model of depth-2 network with \emph{any} width (ranging from finite to continuous).

\begin{definition}
For any measurable functions $\rho:\RR\to\CC$ and $f:\RR^m\to\CC$, put
\begin{align}
    R_\rho[f](\aa,b) := \int_{\RR^m} f(\xx) \overline{\rho(\aa\cdot\xx-b)} \dd \xx, \quad (\aa,b) \in \RR^m \times \RR. \label{eq:ridge}
\end{align}    
We call $R_\rho$ a ridgelet transform.
\end{definition}

The ridgelet transform is known to be a right-inverse operator to $S_\sigma$.
To be precise, the following reconstruction formula holds.
\begin{theorem}[Reconstruction Formula] \label{thm:reconst.ridgelet}
Suppose $\sigma$ and $\rho$ are a tempered distribution ($\calS'$) and a rapid decreasing function ($\calS$) respectively.
There exists a bilinear form $\iiprod{\sigma,\rho}$ such that 
\begin{align}
    S_\sigma \circ R_\rho [f] = \iiprod{\sigma,\rho} f,
\end{align}
for any square integrable function $f \in L^2(\RR^m)$. Further, the bilinear form is given by
\begin{align}
    \iiprod{\sigma,\rho} = \int_{\RR} \sigma^\sharp(\omega) \overline{\rho^\sharp(\omega)} |\omega|^{-m}\dd\omega
\end{align}
where $\sharp$ denotes the 1-dimensional Fourier transform.
\end{theorem}
See \citet[Theorem~6]{Sonoda2021ghost} for the proof.
In particular, according to \citet[Lemma~9]{Sonoda2021ghost}, for any activation function $\sigma$, there always exists $\rho$ satisfying $\iiprod{\sigma,\rho}=1$. 
Here, $\sigma$ being a tempered distribution means that typical activation functions are covered such as ReLU, step function, $\tanh$, gaussian, etc...
We can interpret the reconstruction formula as a universality theorem of continuous neural networks, since for any given data generating function $f$, a network with output weight $\gamma_f = R_\rho[f]$ reproduces $f$ (up to factor $\iiprod{\sigma,\rho}$), i.e. $S[\gamma_f] = f$. In other words, the ridgelet transform indicates how the network parameters should be organized so that the network represents an individual function $f$.

In this study, we showcase a new proof of the reconstruction formula based on the group theoretic arguments, and present a systematic scheme to find the ridgelet transform for a variety of given network architecture based on the symmetry in the data-parameter domain.

\subsection{Irreducible Unitary Representation and Schur's Lemma}
Let $G$ be a locally compact group,
$\calH$ be a nonzero Hilbert space, and
$U(\calH)$ be the group of unitary operators on $\calH$.
For example, any finite group, discrete group, compact group, and finite-dimensional Lie group are locally compact, while an infinite-dimensional Lie group is not locally compact.
A \emph{unitary representation} $\pi$ of $G$ on $\calH$ is a group homomorphism that is continuous with respect to the strong operator topology---that is, 
a map $\pi : G \to U(\calH)$ satisfying $\pi(gh) = \pi(g)\pi(h)$ and $\pi(g^{-1})=\pi(g)^{-1}=\pi(g)^*$, and for any $\psi \in \calH$ map $G \ni g \mapsto \pi(g)[\psi] \in \calH$ is continuous.
Suppose $\calM$ is a closed subspace of $\calH$. $\calM$ is called an \emph{invariant} subspace when $\pi(g)\calM \subset \calM$ for all $g \in G$. Particularly, $\pi$ is called \emph{irreducible} when it does not admit any nontrivial invariant subspace $\calM \neq \{0\}$ nor $\calH$. 

Let $C(\pi)$ be the set of all bounded linear operators $T$ on Hilbert space $\calH$ that commutes with $\pi$, namely $C(\pi) := \{T \in B(\calH) \mid T\pi(g)=\pi(g)T \mbox{ for all } g \in G\}$.
\begin{lemma}[Schur's lemma] \label{lem:schur}
A unitary representation $\pi$ of $G$ is irreducible iff $C(\pi)$ only contains scalar multiples of the identity, i.e., $C(\pi) = \{ c\id \mid c \in \CC \}$ or $\{0\}$.
\end{lemma}
See %
\citet[Theorem~3.5(a)]{Folland2015} for the proof. %

\subsection{Calculus on Locally Compact Group}%
By Haar's theorem, if $G$ is a locally compact group, then there uniquely exist left and right invariant measures $\dd_l g$ and $\dd_r g$, satisfying for any $s \in G$ and $f \in C_c(G)$,
\begin{align*}
    \int_G f(sg) \dd_l g = \int_G f(g) \dd_l g,
    \quad\mbox{and}\quad
    \int_G f(gs) \dd_r g = \int_G f(g) \dd_r g.
\end{align*}

Let $X$ be a $G$-space with transitive left (resp. right) $G$-action $g \cdot x$ (resp. $x \cdot g$) for any $(g,x) \in G \times X$.
Then, we can further induce the left (resp. right) invariant measure $\dd_l x$ (resp. $\dd_r x$) so that for any $f \in C_c(G)$,
\begin{align*}
    \int_X f(x) \dd_l x := \int_G f(g \cdot o) \dd_l g,
    \quad\mbox{resp.}\quad
    \int_X f(x) \dd_r x := \int_G f(o \cdot g) \dd_r g,
\end{align*}
where $o \in G$ is a fixed point called the origin.

\section{Main Results} \label{sec:main} %
We introduce generalized neural networks and generalized ridgelet transforms induced from joint group invariant functions on data-parameter domain, and present a simple group theoretic proof of the reconstruction formula. %

Let $G$ be a locally compact group equipped with a left invariant measure $\dd g$.
Let $X$ and $\Xi$ be $G$-spaces equipped with $G$-invariant measures $\dd x$ and $\dd \xi$, called the data domain and the parameter domain, respectively. Particularly, we call the product space $X \times \Xi$ the \emph{data-parameter} domain (like time-frequency domain).
By abusing notation, we use the same symbol $\cdot$ for the $G$-actions on $X$ and $\Xi$ (e.g., $g \cdot x$ and $g \cdot \xi$).

Let $\pi$ and $\pihat$ be left-regular actions of $G$ on $L^2(X)$ and $L^2(\Xi)$, respectively. Namely, for any $g \in G, f \in L^2(X)$ and $\gamma \in L^2(\Xi)$,
\begin{align}
    \pi_g[f](x) := f(g^{-1} \cdot x), \quad\mbox{and}\quad
    \pihat_g[\gamma](\xi) := \gamma(g^{-1} \cdot \xi).
\end{align}

\begin{figure}
    \centering
    \includegraphics[width=.35\linewidth, trim=91mm 36mm 79mm 40mm, clip]{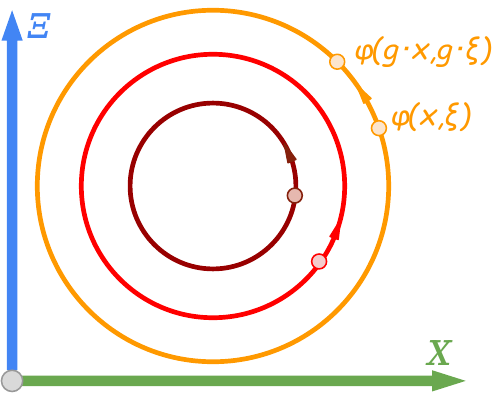}
    \caption{A joint $G$-invariant function $\phi$ is a function on the data-parameter domain $X \times \Xi$ that is constant along each $G$-orbit $G(x,\xi) := \{ (g\cdot x,g \cdot \xi) \mid g \in G\}$.}
    \label{fig:invariant}
\end{figure}

\begin{definition}[Joint $G$-Invariant Function] We say a function $\phi$ on $X \times \Xi$ is joint $G$-invariant when it satisfies
for all $g \in G$ and $(x,\xi) \in X \times \Xi$,
\begin{align}
\phi(g\cdot x,g\cdot \xi) = \phi(x,\xi).
\end{align}
By $\calA$, we symbolize the algebra of all joint $G$-invariant functions.
\end{definition}
Here, $\calA$ is indeed an \emph{algebra} because if $\phi$ and $\psi$ are joint $G$-invariant, then so are $\phi + \psi$ and $\phi \psi$. Namely, $\phi,\psi \in \calA \implies \phi+\psi,\phi \psi \in \calA$. 
As visualized in \reffig{invariant}, a joint $G$-invariant function is constant along each $G$-orbit $\{ (g\cdot x, g \cdot \xi) \mid g \in G \}$. Hence finding a joint $G$-invariant function is not difficult.

\begin{definition}[Generalized Neural Network Induced from Invariant $\phi$]%
For any joint invariant function $\phi \in \calA$ and Borel measure $\gamma$ on $\Xi$, put
\begin{align}
    \nn[\gamma;\phi](x) :=
    \int_\Xi \gamma(\xi) \phi(x,\xi) \dd \xi, \quad x \in X.
\end{align}
We call 
the integral transform $\nn[\bullet;\phi]$ a $\phi$-transform, 
and each individual image $\nn[\gamma;\phi]$ a $\phi$-network for short.
\end{definition}
The $\phi$-network is an extension of the original neural network because when $X=\RR^m, \Xi=\RR^m\times\RR$ and $\phi(\xx,(\aa,b)) := \sigma(\aa\cdot\xx-b)$ with some activation function $\sigma:\RR\to\RR$, it reduces to a fully-connected network
$\int_{\RR^m\times\RR} \gamma(\aa,b) \sigma(\aa\cdot\xx-b)\dd\aa\dd b.$

\begin{definition}[Generalized Ridgelet Transform Induced from Invariant $\phi$]%
For any joint invariant map $\phi \in \calA$ and measurable function $f$ on $X$, put
\begin{align}
    \ridge[f;\phi](\xi) := 
    \int_X f(x) \overline{\phi(x,\xi)} \dd x, \quad \xi \in \Xi.
\end{align}
We call the integral transform $\ridge[\bullet;\phi]$ a $\phi$-ridgelet transform for short.  
\end{definition}
As long as the integrals are convergent, it is the dual operator of $\phi$-transform, since
\begin{align}
    \iprod{ \gamma, \ridge[f;\phi] }_{L^2(\Xi)}
    = \int_{X \times \Xi} \gamma(\xi) \phi(x,\xi) \overline{f(x)} \dd x \dd \xi
    = \iprod{ \nn[\gamma;\phi], f }_{L^2(X)}.
\end{align}

\begin{theorem}
Let $G$ be a locally compact group.
For any joint invariant functions $\phi,\psi \in \calA$, suppose that composite $\nn_\phi \circ \ridge_\psi : L^2(X) \to L^2(X)$ is bounded, and that regular representation $(\pi,L^2(X))$ is irreducible.
Then, there exists a bilinear form $\iiprod{\phi,\psi} \in \CC$
such that for any function $f \in L^2(X)$,
\begin{align}
    \nn_\phi \circ \ridge_\psi [f] = \iiprod{\phi,\psi} f.
\end{align}
\end{theorem}
In other words, the $\psi$-ridgelet transform $\ridge_\psi$ is understood as a group theoretic generalization of the original ridgelet transform, as it is a right inverse operator of $\phi$-transform $\nn_\phi$.

\begin{proof}
We write $\nn[\bullet;\phi]$ as $\nn_\phi$ and $\ridge[\bullet;\phi]$ as $\ridge_\phi$ for short.
By the left-invariances of $\dd x$ and $\psi$, for all $g \in G$, we have
\begin{align}
    \ridge_\psi[\pi_g[f]] (\xi)
    &= \int_X f(g^{-1} \cdot x) \overline{\psi(x,\xi)} \dd x
    &= \iprod{\pi_g [f], \psi(\bullet,\xi)}_{L^2(X)} \notag \\
    &= \int_X f(x) \overline{\psi(g\cdot x,\xi)} \dd x
    &= \iprod{f, \pi_g^*[\psi](\bullet,\xi)}_{L^2(X)}\notag \\
    &= \int_X f(x) \overline{\psi(x,g^{-1}\cdot\xi)} \dd x
    &= \iprod{f, \pihat_g[\psi](\bullet,\xi)}_{L^2(X)}\notag \\
    &= \pihat_g[\ridge_\psi[f]](\xi).
\end{align}
Here, $\pi^*$ denotes the dual representation of $\pi$ with respect to $L^2(X)$-product.
Similarly,
\begin{align}
    \nn_\phi[\pihat_g[\gamma]](x)
    &= \int_\Xi \gamma(g^{-1}\cdot\xi) \phi(x,\xi) \dd \xi
    &= \iprod{ \pihat_g[\gamma], \phi(x,\bullet) }_{L^2(\Xi)}\notag \\
    &= \int_\Xi \gamma(\xi) \phi(x,g\cdot\xi) \dd\xi
    &= \iprod{ \gamma, \pihat_g^*[\phi](x,\bullet) }_{L^2(\Xi)}\notag \\
    &= \int_\Xi \gamma(\xi) \phi(g^{-1}\cdot x,\xi) \dd\xi
    &= \iprod{ \gamma, \pi_g[\phi](x,\bullet) }_{L^2(\Xi)}\notag \\
    &= \pi_g[\nn_\phi[\gamma]](x).
\end{align}
Here, $\pihat^*$ denotes the dual representation of $\pihat$ with respect to $L^2(\Xi)$-product.

As a consequence, $\nn_\phi \circ \ridge_\psi : L^2(X) \to L^2(X)$ commutes with $\pi$ as below
\begin{align}
    \nn_\phi \circ \ridge_\psi \circ \pi_g
    = \nn_\phi \circ \pihat_g \circ \ridge_\psi
    = \pi_g \circ \nn_\phi \circ \ridge_\psi
\end{align}
for all $g \in G$.
Hence by Schur's lemma (\reflem{schur}), there exist a constant $C_{\phi,\psi} \in \CC$ such that 
    $\nn_\phi \circ \ridge_\psi = C_{\phi,\psi} \id_{L^2(X)}$.
By the construction of left-hand side, $C_{\phi,\psi}$ is bilinear in $\phi$ and $\psi$.
\end{proof}

\section{Examples}

\subsection{Original Ridgelet Transform}
This study started from a group theoretic proof of the original reconstruction formula (\refthm{reconst.ridgelet}). The proof is in fact new, thought-provoking and valuable, so we leave it in \refapp{original.proof}. Below is a sketch of the full proof.

\begin{example}
Let $G$ be the affine group $\aff(m) = GL(m) \ltimes \RR^m$, 
$X=\RR^m$ be the data domain with $G$-action
\[g \cdot \xx := L\xx + \tt, \quad g = (L,\tt) \in G, \ \xx \in \RR^m=X\]
and $\Xi=\RR^m\times\RR$ be the parameter domain with \emph{dual $G$-action}
\begin{align}
g \cdot (\aa,b) = (L^{-\top}\aa, b + \tt^\top L^{-\top}\aa), \quad g = (L,\tt) \in G, \ (\aa,b) \in \RR^m\times\RR = \Xi. \label{eq:dual.affine.action}
\end{align}
We can see $\phi(\xx,(\aa,b)) := \sigma( \aa\cdot\xx-b )$ is joint $G$-invariant. In fact,
\begin{align*}
   \phi( g \cdot \xx, g \cdot (\aa,b) )
   = \sigma\left( L^{-\top} \aa \cdot (L\xx+\tt) - (b + \tt^\top L^{-\top} \aa)\right) = \sigma(\aa\cdot\xx-b)
   = \phi( \xx, (\aa,b) ).
\end{align*}
Further, by \reflem{irrep}, the regular representation $\pi_g$ of $G = \aff(m)$ is known to be irreducible. 
Hence we can retain the original neural network and ridgelet transform:
\begin{align*}
\nn[\gamma](\xx) = \int_{\RR^m\times\RR} \gamma(\aa,b) \sigma(\aa\cdot\xx-b)\dd\aa\dd b,
\quad \mbox{and} \quad 
\ridge[f](\aa,b) = \int_{\RR^m} f(\xx) \overline{\rho(\aa\cdot\xx-b)}\dd\xx,
\end{align*}
satisfying $\nn \circ \ridge = \iiprod{\sigma,\rho} \id_{L^2(\RR^m)}$.
\end{example}
Additionally, a geometric interpretation of dual $G$-action \refeq{dual.affine.action} is discussed in \refapp{original.geo}.

\subsection{Deep Ridgelet Transform} %
\citet{sonoda2023deepridge} presented the ridgelet transform for \emph{deep} neural networks. We noticed their network can also be induced from an invariant function. 
In other words, from the group representation theory perspective, function approximation with \emph{any depth} is unified.

\begin{example}
Let $G$ be any locally compact group,
data domain $X$ be any $G$-space, rewriting its $G$-action $g \cdot x$ as $g(x)$ so as to formally identify $g$ with a hidden layer map,
and parameter domain $\Xi$ be the group $G$ itself with dual $G$-action
\begin{align}
g \cdot \xi = \xi g^{-1}.
\end{align}
We can see $\phi(x,\xi) := \psi \circ \xi (x)$ is joint $G$-invariant. In fact,
    \[ \phi(g \cdot x, g \cdot \xi) = \psi \circ (g \cdot \xi)( g \cdot x ) = \psi \circ (\xi \circ g^{-1})( g(x) ) = \psi \circ \xi(x) = \phi(x,\xi)\]
Therefore, assuming that the regular representation $\pi_g = \psi \circ g$ is irreducible on an invariant subspace $\calH$ of $L^2(X)$,
we can retain the formal deep network and deep ridgelet transform:
    \begin{align*}
    \nn[\gamma](x) := \int_\Xi \gamma(\xi) \psi \circ \xi(x) \dd \xi,
    \quad \mbox{and} \quad 
    \ridge[f](\xi) = \int_X f(x) \overline{\psi \circ \xi(x)} \dd x,
    \end{align*}
    satisfying $\nn \circ \ridge = \iiprod{\sigma,\rho} \id_{\calH}$.
\end{example}

\subsection{Voice Transform, or Generalized Wavelet Transform} %

The voice transform is also known as the \emph{Gilmore–Perelomov coherent states} and the \emph{generalized wavelet transform} \citep{Perelomov1986,Ali2014}.
It is well investigated in the research field of \emph{coorbit theory} \citep{Feichtinger1988,Feichtinger1989part1,Feichtinger1989part2}. We refer to \citet{Berge2021} for a quick review of voice transform and coorbit theory.
\begin{definition}
Given a unitary representation $(\pi,\calH)$ of group $G$ on a Hilbert space $\calH$,
the voice transform is defined as
\begin{align}
    V_\phi[f](g) := \iprod{ f, \pi_g[\phi] }_\calH, \quad g \in G, \ f,\phi \in \calH.
\end{align}
\end{definition}
This unifies several integral transforms from the perspective of group theory such as short-time Fourier transform (STFT), wavelet transform   \citep{Grossmann1985i,Grossmann1986ii,Holschneider1998book,Laugesen2002,Gressman2003},
and continuous shearlet transform \citep{Labate2005shearlet,Guo2007,Kutyniok2012}. 

\begin{example}
    Let $G$ be any group,
data domain $X$ be any $G$-space, %
and parameter domain $\Xi$ be the group $G$ itself with dual $G$-action $g \cdot \xi = g \xi$.
We can see $\theta(x,\xi) := \psi(\xi^{-1} \cdot x)$ is joint $G$-invariant. In fact,
    \[ \theta(g \cdot x, g \cdot \xi)
    = \psi( (g \cdot \xi)^{-1} \cdot (g \cdot x) )
    =\psi( \xi^{-1} \cdot x )
     = \theta(x,\xi).\]
Therefore, assuming that the regular representation $\pi_g$ is irreducible,
we can retain a dual voice transform and voice transform:
    \begin{align*}
    \nn[\gamma](x) := \int_\Xi \gamma(\xi) \phi(\xi^{-1}\cdot x) \dd \xi, \quad \mbox{and} \quad 
    \ridge[f](\xi) = \int_X f(x) \psi(\xi^{-1} \cdot x) \dd x,
    \end{align*}
    satisfying $\nn \circ \ridge = \iiprod{\sigma,\rho} \id_{L^2(X)}$. This is a special case of the voice transform when $\calH = L^2(X)$, and $\pi_g[\psi] = \psi(g^{-1}\cdot\bullet)$. 
\end{example}

We note that the voice transform $V_\phi[f](g) := \iprod{f,\pi_g[\phi]}_\calH$ and the $\phi$-ridgelet transform $\ridge_\phi[f](\xi) := \iprod{f, \phi(\bullet,\xi)}_{L^2(X)}$ have common parts, but are different in general.
While the example above and the original wavelet transform $W_\psi[f](b,a) := \int_{\RR} f(x) \psi( (x-b)/a )\dd x/\sqrt{a}$ are simultaneously the voice and ridgelet transforms, 
a ridgelet transform can be a voice transform only when the representation $(\pi,\calH)$ is the regular representation on $L^2(X)$, and a voice transform can be a ridgelet transform only when the parameter domain $\Xi$ is the group $G$ itself and the feature map $\phi$ is generated by $G$-action on a single function $\psi$.
Hence pursuing parallel results for the coorbit theory would be an interesting future work.

\section{Discussion}
We presented a systematic method to induce a generalized neural network and its ridgelet transform, from a joint group invariant function on the data-parameter domain.
Namely, given a joint group invariant function, 
the marginalization of parameter $\xi$ (resp. data $x$) induces the network (resp. the ridgelet transform).
Based on the group theoretic arguments, we demonstrated a simple proof of the reconstruction formula by using Schur's lemma, which implies the universality of the network.
Since conventional universality theorems were shown using functional analytic tools, the group theoretic proof is a new contribution to the approximation theory, connecting geometric deep learning to abstract harmonic analysis.
Further, since the proposed network covers both shallow and deep networks, the group representation theory can offer a unified perspective on function approximation with \emph{any depth}.

In the past, \citet{Sonoda2022gconv,sonoda2022symmetric} have developed the ridgelet transforms for neural networks on manifolds and function spaces using the Fourier transforms on manifolds and function spaces,
and 
proposed a systematic scheme to derive a ridgelet transform for neural networks on a given domain based on the Fourier transform on there.
Compared to our group theoretic method, the Fourier transform method is indirect and requires additional knowledge (not only on the symmetry on the data domain but also) on the Fourier transform on there.
We conjecture that those Fourier-based ridgelet transforms can also be derived in our group-theoretic method.

\acks{The authors are extremely grateful to the three anonymous reviewers for their valuable comments and suggestions, which have helped improve the quality of our manuscript. This work was supported by JSPS KAKENHI 20K03657, JST PRESTO JPMJPR2125, JST CREST JPMJCR2015 and JPMJCR1913, and JST ACTX JPMJAX2004.}

\bibliography{libraryS}

\begin{thebibliography}{47}
\providecommand{\natexlab}[1]{#1}
\providecommand{\url}[1]{\texttt{#1}}
\expandafter\ifx\csname urlstyle\endcsname\relax
  \providecommand{\doi}[1]{doi: #1}\else
  \providecommand{\doi}{doi: \begingroup \urlstyle{rm}\Url}\fi

\bibitem[Ali et~al.(2014)Ali, Antoine, and Gazeau]{Ali2014}
S.~T. Ali, J.-P. Antoine, and J.-P. Gazeau.
\newblock \emph{\href{https://link.springer.com/book/10.1007/978-1-4614-8535-3}{Coherent States, Wavelets, and Their Generalizations}}.
\newblock Theoretical and Mathematical Physics. Springer New York, 2 edition, 2014.

\bibitem[Aslan et~al.(2023)Aslan, Platt, and Sheard]{Aslan2023}
B.~Aslan, D.~Platt, and D.~Sheard.
\newblock \href{https://proceedings.mlr.press/v197/aslan23a.html}{Group invariant machine learning by fundamental domain projections}.
\newblock In \emph{Proceedings of the 1st NeurIPS Workshop on Symmetry and Geometry in Neural Representations}, volume 197 of \emph{Proceedings of Machine Learning Research}, pages 181--218. PMLR, 2023.

\bibitem[Barron(1993)]{Barron1993}
A.~R. Barron.
\newblock \href{http://doi.org/10.1109/18.256500}{{Universal approximation bounds for superpositions of a sigmoidal function}}.
\newblock \emph{IEEE Transactions on Information Theory}, 39\penalty0 (3):\penalty0 930--945, 1993.

\bibitem[Berge(2021)]{Berge2021}
E.~Berge.
\newblock \href{http://doi.org/10.1007/s00041-021-09892-5}{{A Primer on Coorbit Theory}}.
\newblock \emph{Journal of Fourier Analysis and Applications}, 28\penalty0 (2):\penalty0 1--61, 2021.

\bibitem[Bronstein et~al.(2021)Bronstein, Bruna, Cohen, and Veli{\v{c}}kovi{\'{c}}]{Bronstein2021}
M.~M. Bronstein, J.~Bruna, T.~Cohen, and P.~Veli{\v{c}}kovi{\'{c}}.
\newblock \href{http://arxiv.org/abs/2104.13478}{{Geometric Deep Learning: Grids, Groups, Graphs, Geodesics, and Gauges}}.
\newblock \emph{arXiv preprint: 2104.13478}, 2021.

\bibitem[Cand{\`{e}}s(1998)]{Candes.PhD}
E.~J. Cand{\`{e}}s.
\newblock \emph{\href{https://searchworks.stanford.edu/view/9949708}{{Ridgelets: theory and applications}}}.
\newblock PhD thesis, Standford University, 1998.

\bibitem[Cand{\`{e}}s and Donoho(2004)]{Candes2004curvelet}
E.~J. Cand{\`{e}}s and D.~L. Donoho.
\newblock \href{https://doi.org/10.1002/cpa.10116}{New tight frames of curvelets and optimal representations of objects with piecewise C2 singularities}.
\newblock \emph{Communications on Pure and Applied Mathematics}, 57\penalty0 (2):\penalty0 219--266, feb 2004.

\bibitem[Carroll and Dickinson(1989)]{Carroll.Dickinson}
S.~M. Carroll and B.~W. Dickinson.
\newblock \href{http://doi.org/10.1109/IJCNN.1989.118639}{{Construction of neural nets using the Radon transform}}.
\newblock In \emph{International Joint Conference on Neural Networks 1989}, volume~1, pages 607--611. IEEE, 1989.

\bibitem[Chizat and Bach(2018)]{Chizat2018}
L.~Chizat and F.~Bach.
\newblock \href{https://papers.nips.cc/paper/7567-on-the-global-convergence-of-gradient-descent-for-over-parameterized-models-using-optimal-transport/}{{On the Global Convergence of Gradient Descent for Over-parameterized Models using Optimal Transport}}.
\newblock In \emph{Advances in Neural Information Processing Systems 32}, pages 3036--3046, Montreal, BC, 2018.

\bibitem[Cohen et~al.(2019)Cohen, Geiger, and Weiler]{Cohen2019}
T.~S. Cohen, M.~Geiger, and M.~Weiler.
\newblock \href{https://proceedings.neurips.cc/paper/2019/file/b9cfe8b6042cf759dc4c0cccb27a6737-Paper.pdf}{{A General Theory of Equivariant CNNs on Homogeneous Spaces}}.
\newblock In \emph{Advances in Neural Information Processing Systems}, volume~32. Curran Associates, Inc., 2019.

\bibitem[Cybenko(1989)]{Cybenko1989}
G.~Cybenko.
\newblock \href{http://doi.org/10.1007/BF02551274}{{Approximation by superpositions of a sigmoidal function}}.
\newblock \emph{Mathematics of Control, Signals, and Systems (MCSS)}, 2\penalty0 (4):\penalty0 303--314, 1989.

\bibitem[Do and Vetterli(2005)]{Do2005contourlet}
M.~N. Do and M.~Vetterli.
\newblock \href{https://ieeexplore.ieee.org/abstract/document/1532309}{The contourlet transform: an efficient directional multiresolution image representation}.
\newblock \emph{IEEE Transactions on Image Processing}, 14\penalty0 (12):\penalty0 2091--2106, 2005.

\bibitem[Donoho(2002)]{Donoho2002}
D.~L. Donoho.
\newblock \href{http://arxiv.org/abs/math/0212395}{{Emerging applications of geometric multiscale analysis}}.
\newblock \emph{Proceedings of the ICM, Beijing 2002}, I:\penalty0 209--233, 2002.

\bibitem[Feichtinger and Gr{\"{o}}chenig(1988)]{Feichtinger1988}
H.~G. Feichtinger and K.~Gr{\"{o}}chenig.
\newblock \href{https://link.springer.com/chapter/10.1007/BFb0078863}{A unified approach to atomic decompositions via integrable group representations}.
\newblock In \emph{Function Spaces and Applications}, pages 52--73, Berlin, Heidelberg, 1988. Springer Berlin Heidelberg.

\bibitem[Feichtinger and Gr{\"{o}}chenig(1989{\natexlab{a}})]{Feichtinger1989part1}
H.~G. Feichtinger and K.~H. Gr{\"{o}}chenig.
\newblock \href{https://www.sciencedirect.com/science/article/pii/0022123689900554}{Banach spaces related to integrable group representations and their atomic decompositions, I}.
\newblock \emph{Journal of Functional Analysis}, 86\penalty0 (2):\penalty0 307--340, 1989{\natexlab{a}}.

\bibitem[Feichtinger and Gr{\"{o}}chenig(1989{\natexlab{b}})]{Feichtinger1989part2}
H.~G. Feichtinger and K.~H. Gr{\"{o}}chenig.
\newblock \href{https://doi.org/10.1007/BF01308667}{Banach spaces related to integrable group representations and their atomic decompositions. Part II}.
\newblock \emph{Monatshefte f{\"{u}}r Mathematik}, 108\penalty0 (2):\penalty0 129--148, 1989{\natexlab{b}}.

\bibitem[Finzi et~al.(2021)Finzi, Welling, and Wilson]{Finzi2021}
M.~Finzi, M.~Welling, and A.~G.~G. Wilson.
\newblock \href{https://proceedings.mlr.press/v139/finzi21a.html}{A Practical Method for Constructing Equivariant Multilayer Perceptrons for Arbitrary Matrix Groups}.
\newblock In \emph{Proceedings of the 38th International Conference on Machine Learning}, volume 139 of \emph{Proceedings of Machine Learning Research}, pages 3318--3328. PMLR, 2021.

\bibitem[Folland(2015)]{Folland2015}
G.~B. Folland.
\newblock \emph{\href{https://doi.org/10.1201/b19172}{{A Course in Abstract Harmonic Analysis}}}.
\newblock Chapman and Hall/CRC, New York, second edition, 2015.

\bibitem[Funahashi(1989)]{Funahashi1989}
K.-I. Funahashi.
\newblock \href{http://doi.org/10.1016/0893-6080(89)90003-8}{{On the approximate realization of continuous mappings by neural networks}}.
\newblock \emph{Neural Networks}, 2\penalty0 (3):\penalty0 183--192, 1989.

\bibitem[Gressman et~al.(2003)Gressman, Labate, Weiss, and Edward]{Gressman2003}
P.~Gressman, D.~Labate, G.~Weiss, and N.~W. Edward.
\newblock \href{http://doi.org/https://doi.org/10.1016/S1570-579X(03)80036-8}{{8 - Affine, Quasi-Affine and Co-Affine Wavelets}}.
\newblock In \emph{Beyond Wavelets}, volume~10, pages 215--223. Elsevier, 2003.

\bibitem[Grossmann et~al.(1985)Grossmann, Morlet, and Paul]{Grossmann1985i}
A.~Grossmann, J.~Morlet, and T.~Paul.
\newblock \href{https://doi.org/10.1063/1.526761}{Transforms associated to square integrable group representations. I. General results}.
\newblock \emph{Journal of Mathematical Physics}, 26\penalty0 (10):\penalty0 2473--2479, oct 1985.

\bibitem[Grossmann et~al.(1986)Grossmann, Morlet, and Paul]{Grossmann1986ii}
A.~Grossmann, J.~Morlet, and T.~Paul.
\newblock \href{http://www.numdam.org/item/AIHPA_1986__45_3_293_0/}{Transforms associated to square integrable group representations. {II} : examples}.
\newblock \emph{Annales de l'I.H.P. Physique th{\'{e}}orique}, 45\penalty0 (3):\penalty0 293--309, 1986.

\bibitem[Guo and Labate(2007)]{Guo2007}
K.~Guo and D.~Labate.
\newblock \href{https://doi.org/10.1137/060649781}{Optimally Sparse Multidimensional Representation Using Shearlets}.
\newblock \emph{SIAM Journal on Mathematical Analysis}, 39\penalty0 (1):\penalty0 298--318, 2007.

\bibitem[Holschneider(1998)]{Holschneider1998book}
M.~Holschneider.
\newblock \emph{\href{https://global.oup.com/academic/product/wavelets-an-analysis-tool-9780198505211}{{Wavelets: An Analysis Tool}}}.
\newblock Oxford mathematical monographs. Oxford University Press, 1998.

\bibitem[Hornik et~al.(1989)Hornik, Stinchcombe, and White]{Hornik1989}
K.~Hornik, M.~Stinchcombe, and H.~White.
\newblock \href{http://doi.org/10.1016/0893-6080(89)90020-8}{{Multilayer feedforward networks are universal approximators}}.
\newblock \emph{Neural Networks}, 2\penalty0 (5):\penalty0 359--366, 1989.

\bibitem[Koyama et~al.(2023)Koyama, Fukumizu, Hayashi, and Miyato]{Koyama2023}
M.~Koyama, K.~Fukumizu, K.~Hayashi, and T.~Miyato.
\newblock \href{https://arxiv.org/abs/2305.18484}{Neural Fourier Transform: A General Approach to Equivariant Representation Learning}.
\newblock \emph{arXiv preprint: 2305.18484}, 2023.

\bibitem[Kutyniok and Labate(2012)]{Kutyniok2012}
G.~Kutyniok and D.~Labate.
\newblock \emph{\href{http://doi.org/10.1007/978-0-8176-8316-0}{{Shearlets: Multiscale Analysis for Multivariate Data}}}.
\newblock Applied and Numerical Harmonic Analysis. Birkh{\"{a}}user Boston, 1 edition, 2012.

\bibitem[Labate et~al.(2005)Labate, Lim, Kutyniok, and Weiss]{Labate2005shearlet}
D.~Labate, W.-Q. Lim, G.~Kutyniok, and G.~Weiss.
\newblock \href{https://doi.org/10.1117/12.613494}{Sparse multidimensional representation using shearlets}.
\newblock In \emph{Proceedings of Society of Photo-Optical Instrumentation Engineers (SPIE), Wavelets XI}, volume 5914, page 59140U, sep 2005.

\bibitem[Laugesen et~al.(2002)Laugesen, Weaver, Weiss, and Wilson]{Laugesen2002}
R.~S. Laugesen, N.~Weaver, G.~L. Weiss, and E.~N. Wilson.
\newblock \href{http://doi.org/10.1007/BF02930862}{{A characterization of the higher dimensional groups associated with continuous wavelets}}.
\newblock \emph{The Journal of Geometric Analysis}, 12\penalty0 (1):\penalty0 89--102, 2002.

\bibitem[Mallat(2009{\natexlab{a}})]{Mallat2009book}
S.~Mallat.
\newblock \emph{\href{https://www.sciencedirect.com/book/9780123743701/a-wavelet-tour-of-signal-processing}{{A Wavelet Tour of Signal Processing, Third Edition: The Sparse Way}}}.
\newblock Academic Press, 2009{\natexlab{a}}.

\bibitem[Mallat(2009{\natexlab{b}})]{Mallat2009grouplet}
S.~Mallat.
\newblock \href{https://www.sciencedirect.com/science/article/pii/S1063520308000444}{Geometrical grouplets}.
\newblock \emph{Applied and Computational Harmonic Analysis}, 26\penalty0 (2):\penalty0 161--180, 2009{\natexlab{b}}.

\bibitem[Mei et~al.(2018)Mei, Montanari, and Nguyen]{Mei2018}
S.~Mei, A.~Montanari, and P.-M. Nguyen.
\newblock \href{http://doi.org/10.1073/PNAS.1806579115}{{A mean field view of the landscape of two-layer neural networks}}.
\newblock \emph{Proceedings of the National Academy of Sciences}, 115\penalty0 (33):\penalty0 E7665--E7671, 2018.

\bibitem[Miyato et~al.(2022)Miyato, Koyama, and Fukumizu]{miyato2022unsupervised}
T.~Miyato, M.~Koyama, and K.~Fukumizu.
\newblock \href{https://openreview.net/forum?id=7b7iGkuVqlZ}{Unsupervised Learning of Equivariant Structure from Sequences}.
\newblock In \emph{Advances in Neural Information Processing Systems}, 2022.

\bibitem[Murata(1996)]{Murata1996}
N.~Murata.
\newblock \href{http://doi.org/10.1016/0893-6080(96)00000-7}{{An integral representation of functions using three-layered networks and their approximation bounds}}.
\newblock \emph{Neural Networks}, 9\penalty0 (6):\penalty0 947--956, 1996.

\bibitem[Nitanda and Suzuki(2017)]{Nitanda2017}
A.~Nitanda and T.~Suzuki.
\newblock \href{http://arxiv.org/abs/1712.05438}{{Stochastic Particle Gradient Descent for Infinite Ensembles}}.
\newblock \emph{arXiv preprint: 1712.05438}, 2017.

\bibitem[Pennec and Mallat(2005)]{Pennec2005bandelet}
E.~L. Pennec and S.~Mallat.
\newblock \href{https://ieeexplore.ieee.org/document/1407972}{Sparse geometric image representations with bandelets}.
\newblock \emph{IEEE Transactions on Image Processing}, 14\penalty0 (4):\penalty0 423--438, 2005.

\bibitem[Perelomov(1986)]{Perelomov1986}
A.~Perelomov.
\newblock \emph{\href{https://link.springer.com/book/10.1007/978-3-642-61629-7}{Generalized Coherent States and Their Applications}}.
\newblock Theoretical and Mathematical Physics. Springer-Verlag Berlin Heidelberg, 1986.

\bibitem[Ravanbakhsh et~al.(2017)Ravanbakhsh, Schneider, and P{\'{o}}czos]{Ravanbakhsh2017}
S.~Ravanbakhsh, J.~Schneider, and B.~P{\'{o}}czos.
\newblock \href{https://proceedings.mlr.press/v70/ravanbakhsh17a.html}{Equivariance Through Parameter-Sharing}.
\newblock In \emph{Proceedings of the 34th International Conference on Machine Learning}, volume~70 of \emph{Proceedings of Machine Learning Research}, pages 2892--2901. PMLR, 2017.

\bibitem[Rotskoff and Vanden-Eijnden(2018)]{Rotskoff2018}
G.~Rotskoff and E.~Vanden-Eijnden.
\newblock \href{https://proceedings.neurips.cc/paper_files/paper/2018/hash/196f5641aa9dc87067da4ff90fd81e7b-Abstract.html}{{Parameters as interacting particles: long time convergence and asymptotic error scaling of neural networks}}.
\newblock In \emph{Advances in Neural Information Processing Systems 31}, pages 7146--7155, Montreal, BC, 2018.

\bibitem[Sirignano and Spiliopoulos(2020)]{Sirignano2020}
J.~Sirignano and K.~Spiliopoulos.
\newblock \href{http://doi.org/10.1137/18M1192184}{{Mean Field Analysis of Neural Networks: A Law of Large Numbers}}.
\newblock \emph{SIAM Journal on Applied Mathematics}, 80\penalty0 (2):\penalty0 725--752, 2020.

\bibitem[Sonoda et~al.(2021)Sonoda, Ishikawa, and Ikeda]{Sonoda2021ghost}
S.~Sonoda, I.~Ishikawa, and M.~Ikeda.
\newblock \href{http://arxiv.org/abs/2106.04770}{{Ghosts in Neural Networks: Existence, Structure and Role of Infinite-Dimensional Null Space}}.
\newblock \emph{arXiv preprint: 2106.04770}, 2021.

\bibitem[Sonoda et~al.(2022{\natexlab{a}})Sonoda, Ishikawa, and Ikeda]{Sonoda2022gconv}
S.~Sonoda, I.~Ishikawa, and M.~Ikeda.
\newblock \href{http://doi.org/10.48550/arxiv.2205.14819}{{Universality of Group Convolutional Neural Networks Based on Ridgelet Analysis on Groups}}.
\newblock In \emph{Advances in Neural Information Processing Systems 35}, New Orleans, Louisiana, USA, 2022{\natexlab{a}}.

\bibitem[Sonoda et~al.(2022{\natexlab{b}})Sonoda, Ishikawa, and Ikeda]{sonoda2022symmetric}
S.~Sonoda, I.~Ishikawa, and M.~Ikeda.
\newblock \href{https://proceedings.mlr.press/v162/sonoda22a.html}{{Fully-Connected Network on Noncompact Symmetric Space and Ridgelet Transform based on Helgason-Fourier Analysis}}.
\newblock In \emph{Proceedings of the 39th International Conference on Machine Learning}, volume 162, pages 20405--20422, Baltimore, Maryland, USA, 2022{\natexlab{b}}.

\bibitem[Sonoda et~al.(2023)Sonoda, Hashimoto, Ishikawa, and Ikeda]{sonoda2023deepridge}
S.~Sonoda, Y.~Hashimoto, I.~Ishikawa, and M.~Ikeda.
\newblock \href{https://arxiv.org/abs/2310.03529}{Deep Ridgelet Transform: Voice with Koopman Operator Proves Universality of Formal Deep Networks}.
\newblock In \emph{to appear in Proceedings of the 2nd NeurIPS Workshop on Symmetry and Geometry in Neural Representations}, Proceedings of Machine Learning Research. PMLR, 2023.

\bibitem[Starck et~al.(2010)Starck, Murtagh, and Fadili]{Starck2010}
J.-L. Starck, F.~Murtagh, and J.~M. Fadili.
\newblock \href{http://doi.org/10.1017/CBO9780511730344.006}{{The ridgelet and curvelet transforms}}.
\newblock In \emph{Sparse Image and Signal Processing: Wavelets, Curvelets, Morphological Diversity}, pages 89--118. Cambridge University Press, 2010.

\bibitem[Sugiura(1990)]{Sugiura1990book}
M.~Sugiura.
\newblock \emph{\href{https://www.sciencedirect.com/bookseries/north-holland-mathematical-library/vol/44/}{Unitary Representations and Harmonic Analysis: An Introduction}}, volume~44 of \emph{North-Holland Mathematical Library}.
\newblock North-Holland, second edition, 1990.

\bibitem[Suzuki(2020)]{Suzuki2020}
T.~Suzuki.
\newblock \href{https://proceedings.neurips.cc/paper/2020/hash/df1a336b7e0b0cb186de6e66800c43a9-Abstract.html}{{Generalization bound of globally optimal non-convex neural network training: Transportation map estimation by infinite dimensional Langevin dynamics}}.
\newblock In \emph{Advances in Neural Information Processing Systems 33}, pages 19224--19237, 2020.

\end{thebibliography}

\appendix
\section{Group Theoretic Proof for Original Ridgelet Transform} \label{sec:original.proof}
This study started from a group theoretic proof of the original reconstruction formula (\refthm{reconst.ridgelet}). %
The proof is in fact new, thought-provoking and valuable, so we present it here in a self-consistent manner.
A non group theoretic proof by reducing to a Fourier expression is given in \citet[][Theorem~6]{Sonoda2021ghost}.

\subsection{Preliminaries}
We will use the following facts without proofs.

\begin{lemma} \label{lem:bdd.ridge}
Suppose $\sigma$ and $\rho$ are a tempered distribution ($\calS'$) and a Schwartz test function, respectively. Then, $S_\sigma \circ R_\rho : L^2(\RR^m) \to L^2(\RR^m)$ is bounded.
\end{lemma}
See \citet[Lemmas~7 and 8]{Sonoda2021ghost} for the proof.

\begin{lemma} \label{lem:irrep}
The regular representation $\pi$ of the affine group $\aff(\RR^m) := GL(m) \ltimes \RR^m$ on $L^2(\RR^m)$, namely $\pi(g)[f](\xx) := |\det L|^{-1/2} f(L^{-1}(\xx-\tt))$ for any $f \in L^2(\RR^m)$ and $g = (L,\tt) \in \aff(\RR^m)$, is irreducible.
\end{lemma}
See \citet[Theorem~6.42]{Folland2015} for the proof.

\subsection{Proof}
In the following, we identify the group $G$ acting on data domain $\RR^m$ with the affine group $\aff(\RR^m)$, and introduce the so-called twisted dual group action that leaves a function $\theta$ invariant. Then, we see the regular action $\pi$ of $G$ on functions space $L^2(\RR^m)$ commutes with composite $S_\sigma \circ R_\rho$. Hence, by Schur's lemma, $S_\sigma \circ R_\rho$ is a constant multiple of identity, which concludes the assertion of \refthm{reconst.ridgelet}.

\begin{proof}
Let $G$ be the affine group $\aff(\RR^m) = GL(\RR^m) \ltimes \RR^m$.
For any $g = (L,\tt) \in G$, let
\begin{align}
    g \cdot \xx := L \xx + \tt, \quad \xx \in \RR^m
\end{align}
be its action on $\RR^m$, and let
\begin{align}
    \pi(g) [f](\xx)
    &:= |\det L|^{-1/2} f( g^{-1}\cdot\xx ) \notag \\
    &=|\det L|^{-1/2} f( L^{-1}(\xx - \tt) ), \quad f \in L^2(\RR^m)
\end{align}
be its left-regular action on $L^2(\RR^m)$.

Besides, putting
\begin{align}
    \theta((\aa,b), \xx) := \aa\cdot\xx-b, \quad (\aa,b) \in \RR^m\times\RR, \xx \in \RR^m
\end{align}
we define the \emph{twisted dual action} of $G$ on $\RR^m\times\RR$ as
\begin{align}
    g \cdot (\aa,b) := ( L^{-\top}\aa, b + \aa\cdot(L^{-1}\tt) ), \quad (\aa,b) \in \RR^m\times\RR \label{eq:dual.action.ridgelet}
\end{align}
so that the following invariance hold:
\begin{align}
    \theta(g\cdot(\aa,b), g\cdot\xx) = \theta((\aa,b),\xx) = \aa\cdot\xx-b.
\end{align}
To see this, 
use matrix expressions with extended variables
\begin{align}
\theta((\aa,b),\xx)
&= \begin{pmatrix}
    \aa^\top & b
\end{pmatrix}
    \begin{pmatrix}
    I_m & 0 \\0  & -1
    \end{pmatrix}
    \begin{pmatrix}
    \xx \\ 1
    \end{pmatrix}
    =: \ta^\top \tI \tx,\\
    \widetilde{g\cdot\xx}
    &:= 
    \begin{pmatrix}
    g\cdot\xx \\ 1
    \end{pmatrix}
    = 
    \begin{pmatrix}
    L & \tt \\ 0 & 1
    \end{pmatrix}
    \begin{pmatrix}
    \xx \\ 1
    \end{pmatrix}
    =: \tL \tx
\end{align}
and calculate
\begin{align}
    \ta^\top \tI \tx
    = (\ta^\top \tI \tL^{-1} \tI^{-1}) \tI (\tL \tx) = (\tI \tL^{-\top} \tI \ta)^\top \tI (\tL \tx),
\end{align}
which suggests $\widetilde{g\cdot(\aa,b)} := \tI \tL^{-\top} \tI \ta$,
and we have
\begin{align*}
\tI \tL^{-\top} \tI
&= 
    \begin{pmatrix}
    I_m & 0 \\ 0 & -1
    \end{pmatrix}
    \begin{pmatrix}
    L & \tt \\ 0 & 1
    \end{pmatrix}^{-\top}
    \begin{pmatrix}
    I_m & 0 \\ 0 & -1
    \end{pmatrix}\\
    &=
    \begin{pmatrix}
    I_m & 0 \\ 0 & -1
    \end{pmatrix}
    \begin{pmatrix}
    L^{-\top} & 0 \\ -\tt^\top L^{-\top} & 1
    \end{pmatrix}
    \begin{pmatrix}
    I_m & 0 \\0  & -1
    \end{pmatrix}
    =
    \begin{pmatrix}
    L^{-\top} & 0 \\ \tt^\top L^{-\top} & 1
    \end{pmatrix}.
\end{align*}

Further, we define its regular-action by
\begin{align}
    \pihat(g)[\gamma](\aa,b)
    &:= |\det L|^{1/2} \gamma( g^{-1}\cdot(\aa,b) ) \notag \\
    &= |\det L|^{1/2} \gamma( L^\top \aa, b-\aa\cdot\tt ), \quad (\aa,b) \in \RR^m\times\RR.
\end{align}

Then we can see that, for all $g = (L,\tt) \in G$,
\begin{align}
    R_\rho \circ \pi(g) = \pihat(g) \circ R_\rho, \quad \mbox{and} \quad S_\sigma \circ \pihat(g) = \pi(g) \circ S_\sigma. \label{eq:commute.ridgelet}
\end{align}
In fact, at every $g = (L,\tt) \in G$ and $(\aa,b) \in \RR^m\times\RR$, 
\begin{align}
    R_\rho[\pi(g)[f]] (\aa,b)
    &= |\det L|^{-1/2} \int_{\RR^m} f(g^{-1} \cdot \xx) \overline{\rho(\theta((\aa,b),\xx})) \dd \xx \notag\\
    \intertext{by putting $\xx = g \cdot \yy = L\yy+\tt$ with $\dd \xx = |\det L| \dd \yy$,}
    &= |\det L|^{1/2} \int_{\RR^m} f(\yy) \overline{\rho(\theta((\aa,b),g \cdot \yy}))) \dd \yy \notag\\
    &= |\det L|^{1/2} \int_{\RR^m} f(\yy) \overline{\rho(\theta(g^{-1}\cdot(\aa,b), \yy}))) \dd \yy \notag\\
    &= \pihat(g)[R_\rho[f]](\aa,b).
\end{align}
Similarly, at every $g = (L,\tt) \in G$ and $\xx \in \RR^m$,
\begin{align}
    S_\sigma[\pihat(g)[\gamma]](\xx)
    &= |\det L|^{1/2} \int_{\RR^m \times \RR} \gamma(g^{-1}\cdot(\aa,b)) \sigma(\theta((\aa,b),\xx)) \dd \aa \dd b \notag\\
    \intertext{by putting $(\aa,b) := g \cdot (\xxi,\eta) = ( L^{-\top} \xxi, \eta + \xxi\cdot(L^{-1}\tt))$ with $\dd \aa \dd b = |\det L| \dd \xxi \dd \eta$,}
    &= |\det L|^{-1/2} \int_{\RR^m \times \RR} \gamma(\xxi,\eta) \sigma(\theta(g\cdot(\xxi,\eta),\xx)) \dd \xxi \dd \eta \notag\\
    &= |\det L|^{-1/2} \int_{\RR^m \times \RR} \gamma(\xxi,\eta) \sigma(\theta((\xxi,\eta),g^{-1}\cdot\xx)) \dd \xxi \dd \eta \notag\\
    &= \pi(g)[S_\sigma[\gamma]](\xx).
\end{align}   

Hence $S_\sigma \circ R_\rho$ commutes with $\pi(g)$ because 
\begin{align*}
    S_\sigma \circ R_\rho \circ \pi(g) = S_\sigma \circ \pihat(g) \circ R_\rho = \pi(g) \circ S_\sigma \circ R_\rho.
\end{align*}
Since $S_\sigma \circ R_\rho : L^2(\RR^m) \to L^2(\RR^m)$ is bounded (\reflem{bdd.ridge}),
and $(\pi,L^2(\RR^m))$ is an irreducible unitary representation of $G$ (\reflem{irrep}), Schur's lemma (\reflem{schur}) yields that there exist a constant $C_{\sigma,\rho} \in \CC$ such that 
\begin{align}
    S_\sigma \circ R_\rho [f] = C_{\sigma,\rho}f
\end{align}
for all $f \in L^2(\RR^m)$. 

Finally, by directly computing the left-hand-side, namely $S_\sigma \circ R_\rho[f]$,
we can verify that the constant $C_{\sigma,\rho}$ is given by
\begin{align}
    C_{\sigma,\rho} = \iiprod{\sigma,\rho} := (2\pi)^{m-1} \int_\RR \sigma^\sharp(\omega)\overline{\rho^\sharp(\omega)} |\omega|^{-m} \dd\omega.
\end{align}
\end{proof}

\newcommand{\dist}[1]{d_E(#1)}
\section{Geometric Interpretation of Dual Action for Original Ridgelet Transform} \label{sec:original.geo}
We explain a geometric interpretation of the dual action \refeq{dual.action.ridgelet} in the previous section. 
We note that in general $\theta$ does not require any geometric interpretation as long as it is joint group invariant on data-parameter domain.

For each $(\aa,b) \in \RR^m\times\RR$, put $\xi(\aa,b) := \{ \xx \in \RR^m \mid \aa\cdot\xx-b=0 \}$. Then it is a hyperplane in $\RR^m$ through point $\xx_0 = b\aa/|\aa|^2$ with normal vector $\uu := \aa/|\aa|$.

\begin{figure}[h]
    \centering
    \includegraphics[width=.4\linewidth]{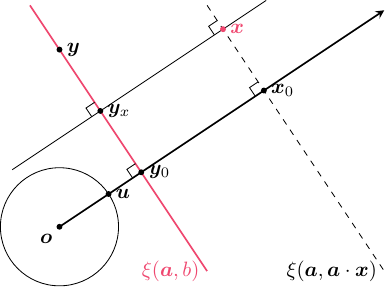}
    \caption{The invariant $\phi((\aa,b), \xx) = \sigma(\aa\cdot\xx-b)$ is the euclidean distance between point $\xx$ and hyperplane $\xi(\aa,b)$ followed by scaling and nonlinearity $\sigma$}
    \label{fig:interpret.ridge}
\end{figure}

For any point $\yy$ in the hyperplane $\xi(\aa,b)$,
by definition $\aa\cdot\yy=b$, thus
\begin{align}
    \aa\cdot\xx-b &= \aa\cdot(\xx-\yy).
\intertext{But this means $\aa\cdot\xx-b$ is a scaled distance between point $\xx$ and hyperplane $\xi(\aa,b)$,}
    &= |\aa| \dist{\xx, \xi(\aa,b)},
\intertext{and further a scaled distance between hyperplanes $\xi(\aa,\aa\cdot\xx)$ through $\xx$ with normal $\aa/|\aa|$ and $\xi(\aa,b)$,}
    &= |\aa| \dist{\xi(\aa,\aa\cdot\xx), \xi(\aa,b)}.
\end{align}

Now, we can interpret the invariant $\theta( (\aa,b), \xx ) := \aa\cdot\xx-b$ in a geometric manner, that is, it is the distance between point and hyperplane, or between hyperplanes.
We note that we can regard entire $\sigma(\aa\cdot\xx-b)$---the distance modulated by both scaling and nonlinearity---as the invariant, say $\phi$.

Furthermore, the dual action $g \cdot (\aa,b)$ is understood as a parallel translation of hyperplane $\xi(\aa,b)$ to $\xi(g \cdot (\aa,b))$ so as to leave the scaled distance $\theta$ invariant, namely 
\begin{align}
\dist{g \cdot \xx, g \cdot \xi(\aa,b)} = \dist{\xx, \xi(\aa,b)}.
\end{align}
Indeed, for any $g = (L,\tt) \in G$,
\begin{align*}
    g \cdot \xi(\aa,b)
    &= \{ g \cdot \xx \mid \aa\cdot\xx-b=0 \}\\
    &= \{ \yy \mid \aa\cdot(g^{-1}\cdot\yy)-b=0 \} \tag{by letting $\yy=g\cdot\xx$}\\
    &= \{ \yy \mid (L^{-\top})\cdot\yy - (b + \aa\cdot (L^{-1}\tt))=0 \}\\
    &= \xi( g \cdot (\aa,b) ),
\end{align*}
meaning that the hyperplane with parameter $(\aa,b)$ translated by $g$ is identical to the hyperplane with parameter $g \cdot (\aa,b)$.

To summarize, in the case of fully-connected neural network (and its corresponding ridgelet transform), the invariant is 
a modulated distance $\sigma(\aa\cdot\xx-b)$, and the dual action is the parallel translation of hyperplane so as to keep the distance invariant. Further, from this geometric perspective, we can rewrite the fully-connected neural network in a geometric manner as
\begin{align}
    S[\gamma](\xx) := \int_{\RR \times \Xi} \gamma( \xi ) \sigma( a \dist{\xx, \xi} ) \dd a \dd \xi,
\end{align}
where $a \in \RR$ denotes signed scale and $\Xi$ denotes the space of all hyperplanes (not always through the origin). Since each hyperplane is parametrized by normal vectors $\uu \in \SS^{m-1}$ and distance $p \ge 0$ from the origin, we can induce the product of spherical measure $\dd\uu$ and Lebesgue measure $\dd p$ as a measure $\dd\xi$ on the space $\Xi$ of hyperplanes.

\end{document}